# Multi-perspective monitoring of wildlife and human activities from camera traps and drones with deep learning models

Hao Chen[a], Fang Qiu[a*], Li An[b,c], Douglas Stow[d], Eve Bohnett[b,e], Haitao Lyu[a], Shuang Tian[a]

a. Geospatial Information Science, the University of Texas at Dallas, Richardson, TX 75080, United States

b. College of Forestry, Wildlife and Environment, Auburn University, Auburn, AL 36849, United States

c. International Center for Climate and Global Change Research, College of Forestry, Wildlife and Environment, Auburn University, 602 Duncan Drive, Auburn, AL 36849, United States

d. Department of Geography, San Diego State University, San Diego, CA 92182, USA

e. Department of Landscape Architecture, University of Florida, Gainesville, FL 32611, United States

Corresponding author: Fang Qiu, Geospatial Information Sciences, University of Texas at Dallas, 800 West Campbell Road, Richardson, TX, 75080, United States, ffqiu@utdallas.edu

**Context**: Wildlife and human activities are key components of landscape systems. Understanding their spatial distribution is essential for evaluating human–wildlife interactions and informing effective conservation planning.

**Objectives**: Multi-perspective monitoring of wildlife and human activities by combining camera traps and drone imagery. Capturing the spatial patterns of their distributions, which allows the identification of the overlap of their activity zones and the assessment of the degree of human-wildlife conflict.

**Methods**: The study was conducted in Chitwan National Park (CNP), Nepal, and adjacent regions. Images collected by visible and near-infrared camera traps and thermal infrared drones from February to July 2022 were processed to create training and testing datasets, which were used to build deep learning models to automatic identify wildlife and human activities. Drone-collected thermal imagery was used for detecting targets to provide a multiple monitoring perspective. Spatial pattern analysis was performed to identify animal and resident activity hotspots and delineation potential human–wildlife conflict zones.

**Results**: Among the deep learning models tested, YOLOv11s achieved the highest performance with a precision of 96.2%, recall of 92.3%, mAP50 of 96.7%, and mAP50-90 of 81.3, making it the most effective for detecting objects in camera trap imagery. Drone-based thermal imagery, analyzed with an enhanced Faster R-CNN model, added a complementary aerial viewpoint for camera trap detections. Spatial pattern analysis identified clear hotspots for both wildlife and human activities and their overlapping patterns within certain areas in the CNP and buffer zones indicating potential conflict.

**Conclusions**: This study reveals human–wildlife conflicts within the conserved landscape. Integrating multi-perspective monitoring with automated object detection enhances wildlife surveillance and landscape management.

## Introduction

Human activities frequently change or destroy the habitats of the wildlife, with illegal hunting and killing being the most detrimental impacts (Ikeda et al., 2022; Lee et al., 2024; Lewis et al., 2021). Wildlife in the natural environment will interact with humans to a certain extent, thus affecting the human population, land use, land cover, etc. (Soulsbury et al., 2015). Human-wildlife interactions vary in frequency and can range from positive to negative, with negative interactions often referred to as human-wildlife conflict (Nyhus, 2016). Monitoring wildlife and human activities is essential for understanding human-wildlife interactions, which play a crucial role in effective wildlife conservation and sustainable ecological landscape management.

Camera traps are pointed-based detectors focusing on a specific point in space, which have been widely employed for wildlife observation, contributing significantly to advancements in understanding human-wildlife interaction, biodiversity richness, and the conservation of endangered species (Caravaggi et al., 2017; Linkie and Martin, 2011; Pierini, et al., 2024; Rowcliffe et al., 2008). Camera traps are often equipped with visible and near-infrared sensors, which automatically gather data upon detecting any motion within their range. Compared to traditional observation methods, camera traps provide continuous 24/7 surveillance and allow for non-invasive monitoring, which minimizes harm, disturbance, or stress to animals and their habitats. Additionally, camera traps can gather rich environmental data, including GPS coordinates, collecting time, and temperature (Tanwar et al., 2009; Jasiulionis et al., 2023; Palmer et al., 2018). However, analyzing the millions of pictures and videos captured by camera traps presents significant human and financial challenges. Conventional manual analysis techniques are not only time-consuming but also struggle with obstructed wildlife, fast-moving or blurry animals, partial body visibility, weak light conditions, and multiple targets overlapping each other (Fig. 1).

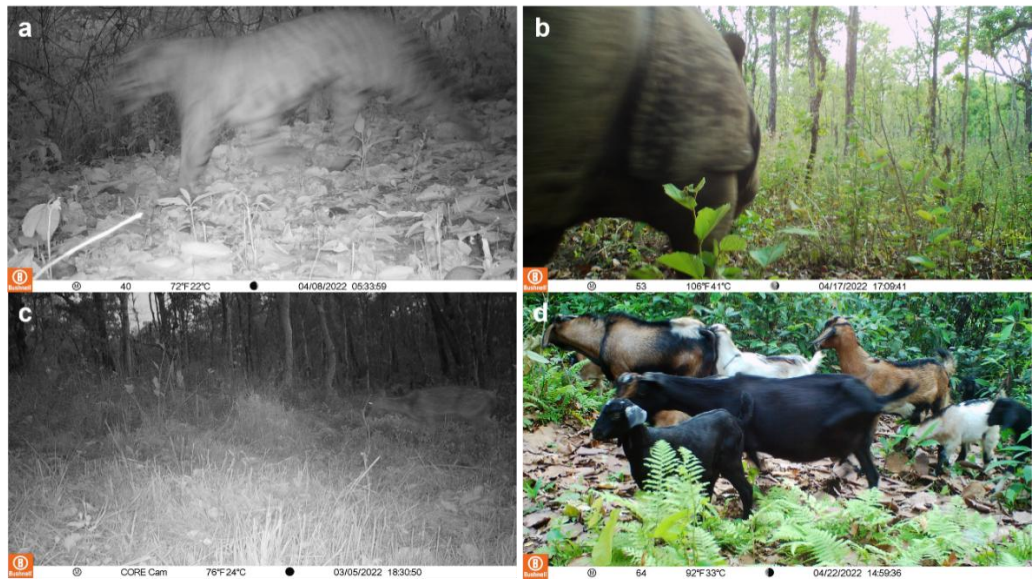

**Fig. 1** Camera trapping images with (a) a fast-moving tiger, (b) a partially visible rhino, (c) a spotted deer under weak light, and (d) multiple overlapping domestic goats.

Drone imagery has emerged as a powerful tool for aerial wildlife monitoring, providing expansive coverage and enabling access to remote or difficult-to-reach areas that are challenging for traditional ground surveys (Corcoran et al., 2021; Koger et al., 2023). Unlike stationary camera traps, drones offer flexibility of movement, allowing researchers to actively survey diverse landscapes and adapt flight paths as needed, thereby enhancing animal tracking efforts. Among the various sensor types, thermal sensors detect targets by capturing variations in infrared radiation emitted due to temperature differences between targets and their surrounding environment, enabling clear, high-contrast differentiation between living organisms and background elements (Lee et al., 2021). Drones quipped with thermal sensors have demonstrated superior capabilities over standard optical systems, particularly in detecting wildlife obscured by vegetation, operating under low-light conditions, or navigating complex environmental conditions (Beaver et al., 2020). Moreover, the integration of drone imagery with automated detection algorithms holds significant potential for advancing fully automated monitoring of ecological communities, contributing to more accurate wildlife population estimation. (Besson et al., 2022; Povlsen et al., 2023; Lyu et al., 2024).

In recent years, deep learning methods have performed advantages in both efficiency and accuracy in target detection. There are mainly two types of deep learning object detectors (Fig. 2). Two-stage detectors, including Faster RCNN (Region-based Convolutional Neural Networks) (Ren et al., 2016), R-FCN (Region-based Fully Convolutional Networks) (Dai et al., 2016), and Mask R-CNN (He et al., 2017), divide the detection process into two stages, region proposal and classification. Two stage detectors usually reach better accuracy but are slower than one-stage detectors. On the other hand, one-stage detectors, such as YOLO (You Only Look Once) (Redmon et al., 2016), SSD (Single Shot Multi-Box Detector) (Liu et al., 2016), and RetinaNet (Lin et al., 2017) directly predict bounding boxes and class probabilities in a single step. These models typically result in a trade-off, achieving higher processing speeds at the expense of some accuracy.

Utilizing deep learning techniques for target detection in camera trap imagery enables automation, enhances efficiency and accuracy, and substantially reduces costs (Christin et al., 2019; Whytock et al., 2021; Petso et al., 2022). Yu et al. (2013) analyzed the target detection using SVM (support vector machine) classifier with a camera trap dataset achieving an average classification accuracy of 82%. Villa et al. (2017) addressed automatic species recognition on the Snapshot Serengeti (SS) dataset, analyzing the species identification task using Very Deep Convolutional Neural Networks. Tabak et al. (2018) trained CNN-based models to classify wildlife species from camera trap images obtained from five states across the United States with 3,367,383 images and achieved 98% accuracy. The Faster R-CNN detection model tested more accurately when applied in wildlife detection with camera trap dataset (Carl et al., 2020; Simoes et.al. 2023). However, detection efficiency remains a major challenge for two-stage detectors.

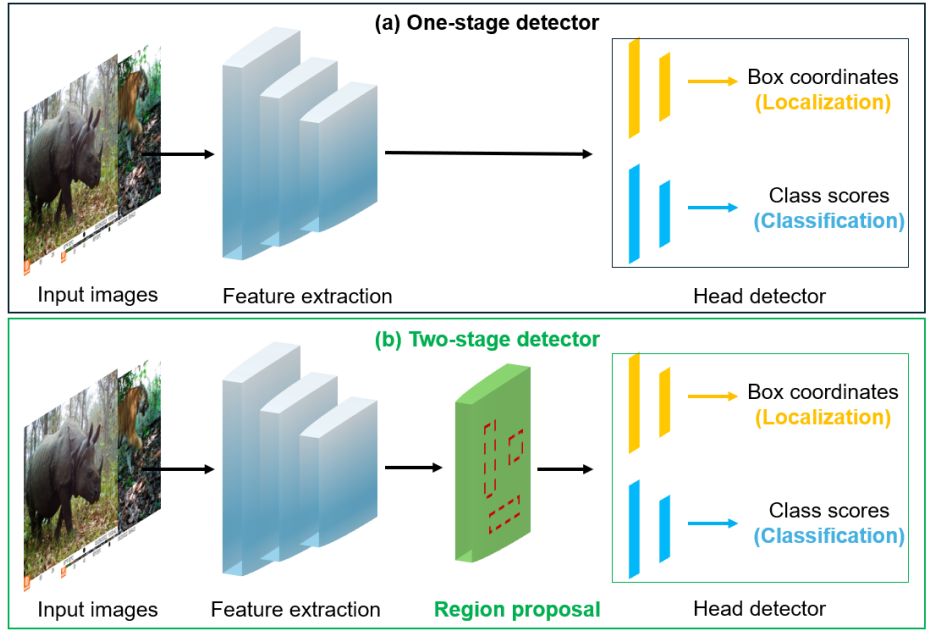

**Fig. 2** Two types of deep learning object detectors: (a) one-stage detector and (b) two-stage detector.

As a one-stage detector, YOLO models are designed for real-time detection, offering rapid processing speeds and high efficiency while maintaining strong detection accuracy (Diwan et al., 2023; Ferrante et al., 2024). Bjerge et al. (2023) used YOLO models to detect and classify small insects, which demonstrated that YOLOv5 has a better performance than YOLOv3. Leonid et al. (2023) proposed to detect wild animals using YOLOv4 model and achieved 94% accuracy with camera trap images. Mamalis et al. (2023) demonstrated YOLOv5 model can detect olive trees and predict their status with UAV imagery. Xie et.al. (2023) improved the YOLOv5 model to recognize big mamma species with airborne thermal imagery and got 96% recognition accuracy. Povlsen et.al. (2024) trained YOLOv5 model to detect hare and roe deer in thermal aerial video footage and provided a conceptual algorithm for implementing real-time object detection in uncrewed aircraft systems. To effectively study human-wildlife interactions, high efficiency detection of camera trap imagery is essential, along with spatial analysis of the detected wildlife and human activities.

The goal of this study is to integrate camera traps and drone-based monitoring to enable multi-scale assessment of wildlife and human activities within a conservation landscape. Specifically, we aim to develop a deep learning dataset for wildlife and human activities, apply object detection in real-word national park scenarios, and conduct spatial analyses to interpret the distribution patterns. The key contributions of this research are as follows: (1) Integration of multi-scale landscape monitoring with automated object detection through using camera traps and drone imagery, (2) development of a camera trapping dataset from a protected landscape, containing both wildlife and human activities, to support deep learning applications, (3) training and evaluation of various deep learning models on this dataset to access their performance in automated detection tasks, and (4) spatial analysis of wildlife and human activities, identify potential zones of human-wildlife conflicts, and inform landscape-scale conservation strategies.

## Methods

Study area

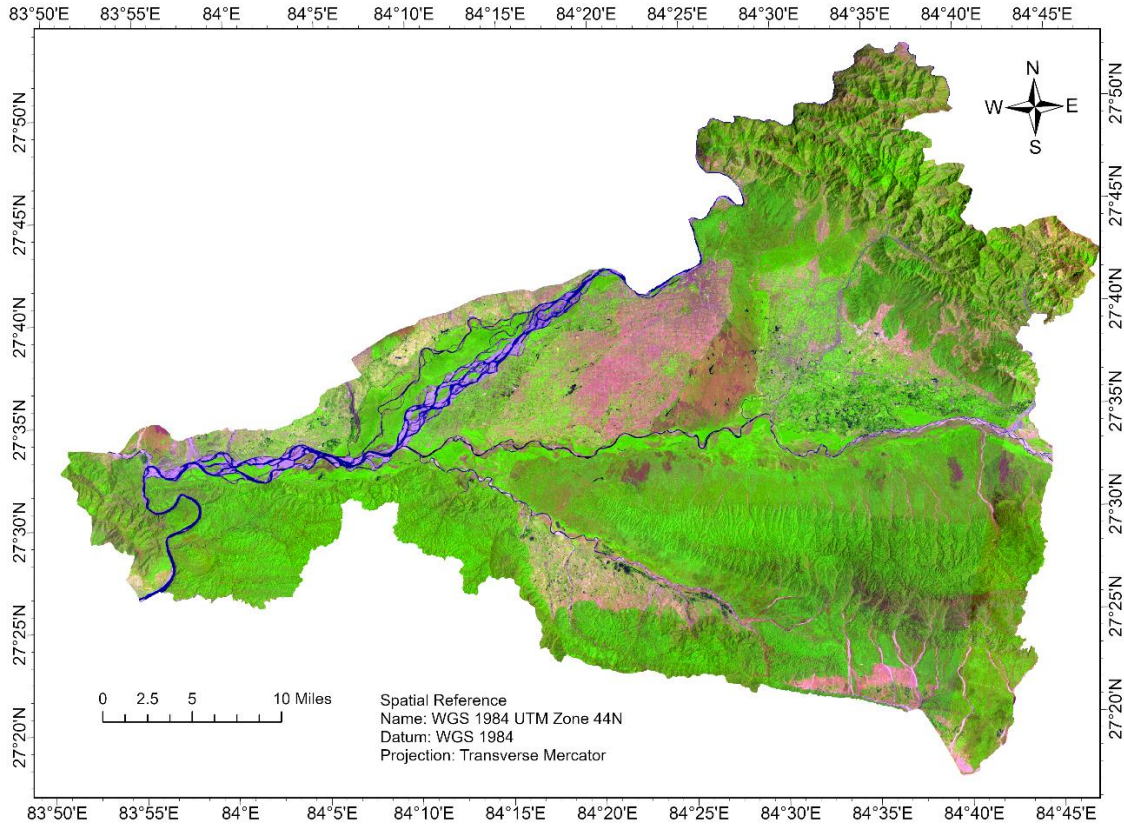

**Fig. 3** Chitwan National Park and its surrounding regions (False color image of Landsat 8 acquisition on April 5, 2022. Red: Band 6; Green: band 5; Blue: band 4).

We conducted the study in Chitwan National Park (CNP) and its surrounding regions (Fig. 3). Established in 1973 as the first national park in Nepal, CNP was listed as a World Heritage Site in 1984, considering its unique ecosystems of international significance (Mishra, 1982). It covers an area of 932 km$^2$ in south central Nepal and is a part of the Chitwan-Parsa-Valmiki Tiger Conservation Landscape in which reducing human-wildlife conflict is supposed to be one of the key outcomes (Bhattarai et al., 2019; Bhatt et.al, 2023). Surrounding the CNP is a buffer zone encompassing 750 km$^2$ of forests and private lands, including cultivated lands. A human population of over 273,977 are living in the buffer zone of CNP (Dangol et al., 2020). Local people depend on park resources for their livelihood, collecting wood, grass and various herbs, either legally or illegally.

The national park consists of a diversity of ecosystems, including hills, lakes, flood plains, and rivers, which nurture endangered great rhinos (One Horned Rhinoceros), tigers (Royal Bengal Tiger), crocodiles (Gharial Crocodile), and other endemic species (Dai et.al., 2020; UNESCO, 2019). The climate in the park is subtropical, primarily influenced by the southeast monsoon. From March to June, temperatures can reach as high as 43 ℃. The monsoon season typically follows, lasting from late June until September. The mean annual rainfall of the park is about 2,150 mm.

Data collection

Camera trapping data were collected throughout the study area to support the training and detection of deep learning models. The images were captured with motion-sensitive camera traps (Bushnell Core DS-4K), which produce motion-activated 32MP photos and 4K 30 fps videos. Featuring one camera sensor optimized for daylight photography and another designed for sharp photos after dark, the camera traps will capture images and videos with the date, time, temperature, and GPS coordinates stamped on the photos (Fig. 4). The camera systems are strategically positioned at sites exhibiting signs of wildlife and human activities, with GPS coordinates recorded for each trap location.

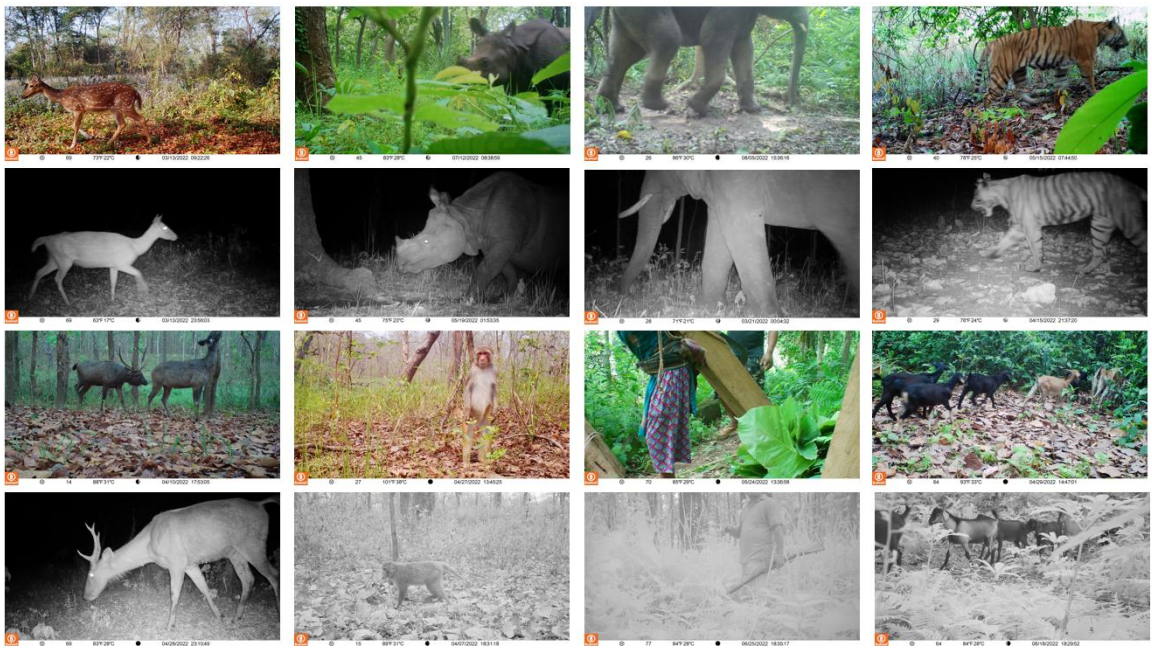

**Fig. 4** Camera trapping images captured day and night.

In the experimental area, a total of 80 camera traps were deployed for data collection from February to July 2022, with 20 traps placed within the CNP, 20 in the buffer zone, and 40 in the North of Barandabhar corridor forest and community forest (Fig. 5). Barandabhar corridor forest (BCF) connects Terai Arc Landscape (TAL) and Chitwan Annapurna Landscape (CAL), which is the only remaining natural forest that allows the endangered one-horned rhino and Royal Bengal tiger access to refuge at higher altitudes during monsoon seasons (Aryal et al., 2012; Lamichhane et al., 2018). Community forest management program is one of the most prominent forest managements, aimed at conserving and restoring forest cover while enhancing the social and economic benefits to local communities. Local communities are allowed to establish community forest user groups, which are given the right to manage, protect, and utilize forest resources sustainably (Smith et al., 2023). The camera traps are maintained by trained professionals and transmit data regularly.

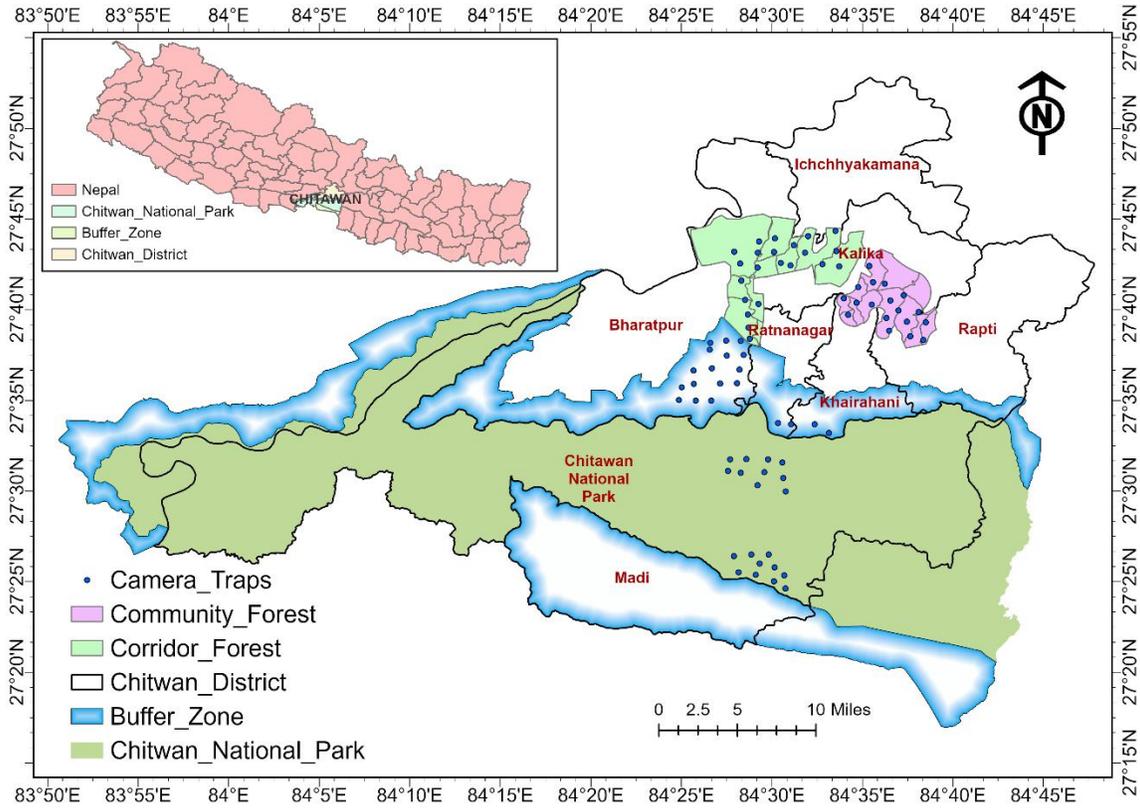

**Fig. 5** Location of camera traps in the research area.

To reflect human activities, images including human and domestic goats are collected in the dataset. Images of wildlife, including spotted deer, sambar deer, rhesus macaques, rhinos, tigers, and elephants, are also collected. Deer and rhesus macaques are the most common wildlife in the national park, while rhinos and tigers are the most endangered wildlife in the world. Elephants are the world's largest terrestrial mammal and are reported to destroy crops, buildings, and even kill people (Dangol et al., 2020). The Roboflow platform was utilized to create datasets, which allowed us to annotate images accurately and export the results into YOLO model efficiently (Roboflow, 2024).

**Table 1.** Wildlife and human activities dataset.

| Class | Domestic Goat | Tiger | Rhino | Rhesus macaques | Elephant | Sabar Deer | Spotted Deer | Human |
|---|---|---|---|---|---|---|---|---|
| Image | 1120 | 1211 | 2733 | 2220 | 1776 | 3639 | 2145 | 1842 |
| Annotate | 3042 | 1211 | 2805 | 2997 | 1884 | 3931 | 4996 | 2081 |

To maintain class balance within the dataset, underrepresented classes such as tiger, rhino, and elephant were augmented using random rotation using Roboflow. The final distribution of classes in the dataset is presented in Table 1. The dataset was resized to 640*640 and split into three portions, allocating 70% for training, 20% for validation, and 10% for testing. The proposed camera trapping dataset facilitates the use of automated object detection models in a national park to identify, locate, count, and map the distribution the wildlife and human activities.

Drone images were collected across the study area between February and July 2022 using a DJI Mavic 2 Enterprise, equipped with both thermal and optical visual cameras. This system simultaneously captured infrared and visible RGB images through a fully stabilized 3-axis gimbal. The RGB images have a resolution of 4000 x 3000 pixels, whereas the thermal images are captured at a resolution of 640 x 512. In total, 22,478 thermal images were captured in the study area. For target detection, we utilized the dataset published from deer survey study (Lyu et al., 2024), which included 2,278 thermal images with 13,509 deer instance annotations. The dataset was divided into training, validation, and testing sets using a 75:15:15 split.

Deep learning models

YOLO models

YOLOv5 is one of the most widely used models in the family of YOLO models, known for its efficiency and flexibility. It is primarily an anchor-based model but can support anchor-free approaches. The architecture comprises the cross-stage partial (CSP) Darknet backbone for enhancing the feature extraction, the spatial pyramid pooling (SPP) and path aggregation network (PANet) in the neck for multi-scale feature fusion, and three YOLO detection heads to predict small, medium, and large targets. YOLOv5 offers five different network sizes: nano (n), small (s), medium (m), large (l), and extra-large (x). These sizes differ in complexity and scale, achieved by varying the depth and width of the model.

During model training, YOLOv5 calculates complete intersection over union (CIOU) to compute the location loss and binary cross entropy (BCE) for the classification loss. With non-maximum suppression (NMS), the model enhances detection accuracy and effectively addresses issues of target occlusion. A weighted loss function combining classification loss, objection loss, and location loss, allows the model to learn from training data and achieve a low prediction error rate when making predictions. Additionally, YOLOv5 makes three kinds of augmentations in each training batch, including scaling, color space adjustment, and Mosaic, to improve small object detection and reduce overfitting.

YOLOv11 (Khanam and Hussain, 2024) is the latest version in the YOLO series. Its backbone incorporates a spatial pyramid pooling-fast (SPPF) block and a new cross stage partial with spatial attention (C2PSA) block, designed to more effectively emphasize critical regions in an image. In the neck, the traditional C2f block is replaced with the more efficient C3k2 block, enhancing the overall performance of the feature aggregation. The head integrates the C3k2 block along with Convolution-BatchNorm-SiLU (CBS) blocks, which collectively refine the feature map by focusing on essential features, stabilizing and normalizing the data flow, and improving the non-linear prediction performance.

In this study, we utilized the latest version of YOLOv5 model, specifically YOLOv5n, YOLOv5s, and YOLOv5m, along with YOLOv11n, YOLOv11s, and YOLOv11m. The models were trained using an input image size of 640, a batch size of 32, and 150 epochs. To improve robustness and minimize overfitting, image augmentation techniques, including hue adjustment, saturation, translation, scaling, and mosaic, were incorporated into the hyperparameter configuration. All training and evaluation processes were conducted on a dedicated NVIDIA RTX 4070 GPU.

Faster R-CNN based model

We utilized the enhanced Faster R-CNN model developed in Lyu et al., (2024) for deer detection for all the collected thermal images. In this model, feature pyramid network and residual networks were used to construct multi-scale feature maps with different resolutions. Moreover, customized anchor boxes that match deer of different sizes in the drone thermal images were adopted to improve the precision of small object detection. By comparing different Faster R-CNN detection networks using COCO evaluation matrix, the integration of Faster R-CNN, FPN, and ResNet18 is proved to be the best, achieving an Average precision score of 91.6% for all deer objects.

For wildlife detection, the complete set of 22,478 thermal images was input into the best-performing Faster R-CNN model. The resulting detection outputs were then used to estimate deer density within the study area, providing a spatially explicit measure of abundance. In addition, these drone-based detection results served as an independent reference to validate the findings derived from camera trap data, allowing for cross-verification between the two monitoring methods and enhancing the reliability of the density estimation.

Evaluation metrics

This study evaluates the performance of YOLO models with four primary metrics, including precision, recall, mAP50 (mean Average Precision), and mAP50-95. Precision and recall were calculated with Equations 1 and 2, where TP, FP, and FN denote the true positive, false positive, and false negative, respectively. Precision evaluates the accuracy of the positive predictions made by the model, indicating how many of the predicted positives are correct. Recall, also referred to as sensitivity, assesses the ability of the model to identify all the relevant instances within the dataset.

$$Precision = \frac{TP}{TP+FP} \tag{1}$$

$$Recall = \frac{TP}{TP+FN} \tag{2}$$

The mAP50 represents the average precision calculated across all recall levels for an IoU (Intersection over Union) threshold greater than 0.50, whereas mAP50-95 indicates the average precision across all recall levels for IoU thresholds ranging from 0.50 to 0.95.

Mapping wildlife and human activities

Hotspots analysis

Hotspots where wildlife and human activities intersect are especially valuable for wildlife conservation efforts (Ahmad et al., 2024; Bagheriyan et al., 2023). Identifying hotspots can assist conservation organizations and governments in formulating protection strategies and optimizing buffer zones. In this paper, we employed methods based on Kernel Density Estimation (KDE) to identify hotspots where wildlife and human activities overlap, using results from camera trapping detection. Human activities were represented by detections of people and domestic goats. Wildlife detections, such as tigers and rhinos, were aggregated to represent overall wildlife presence. However, deer and rhesus macaques, which are prevalent in national parks, and elephants, frequently utilized for transporting humans in Nepal, were excluded from this grouping (Szydlowski, 2022).

The Kernel Density spatial analysis tool and heat map in ArcGIS Pro were utilized to calculate the density by applying a Kernel function, which generate a smoothly tapered surface around each point. The frequency of occurrences of each specific class in the camera trapping images was used as the population field for spreading across the landscape to create a continuous surface (Rovero et al., 2009). The detection results from drone thermal images were employed to calculate the spatial point density.

Clustering the camera traps

Clustering is an unsupervised learning method used in pattern recognition and machine learning to identify subgroups, in which the observations within each group are quite similar to each other, while observations in different groups show significant dissimilarity. Hierarchical clustering is a commonly utilized technique for grouping observations, following a bottom-up approach that generates a tree-based representation of the observations as a dendrogram (Murtagh and Contreras, 2012). By analyzing the dendrograms, branches can be merged progressively as we move up the tree. The dissimilarity measure between each pair of observations is typically measured using Euclidean distance or correlation-based distance. Correlation-based distance identifies two observations as similar if their features exhibit strong correlations, even though the observed values may be significantly different in terms of Euclidean distance (James et al. 2023). This approach emphasizes the pattern or shape of the observation profiles rather than their absolute magnitudes.

In this paper, the detection frequencies of wildlife and human activities were independently aggregated and analyzed using correlation-based hierarchical clustering to categorize the camera trap locations. The algorithm was executed using Python scripts.

## Results

Model training

**Table 2**. Training results of the YOLO models.

| Models | Precision | Recall | mAP50 | mAP50-95 | GFLOPs | Parameters (M) |
|--------|-----------|--------|-------|----------|--------|----------------|
| YOLOv5n | 0.936 | 0.893 | 0.951 | 0.723 | 4.2 | 1.77 |
| YOLOv5s | 0.952 | 0.925 | 0.963 | 0.761 | 15.8 | 7.03 |
| YOLOv5m | 0.955 | <u>0.935</u> | 0.962 | 0.786 | 47.9 | 20.88 |
| YOLOv11n | 0.950 | 0.909 | 0.961 | 0.794 | 6.3 | 2.58 |
| **YOLOv11s** | <u>0.962</u> | 0.923 | <u>0.967</u> | 0.813 | 21.3 | 9.41 |
| YOLOv11m | 0.958 | 0.931 | 0.965 | <u>0.819</u> | 67.7 | 20.03 |

The comprehensive training results of the YOLO models using the camera trapping dataset are presented in Table 2. The computational complexity of each model is quantified using GFLOPs (Giga Floating-Point Operations per Second), which represents the total number of floating-point operations performed. Generally, higher GFLOPs indicate a more computationally intensive and complex model. Additionally, the model parameters are measured in millions (M), with an increasing number of parameters correlating to enhanced performance across both YOLOv5 and YOLOv11 architectures.

A comparative analysis reveals that YOLOv11 consistently outperforms YOLOv5 at an equivalent network size. For instance, YOLOv11n achieves higher performance with a precision of 95.0%, recall of 90.9%, mAP50 of 96.1%, and mAP50-95 of 79.4%, surpassing YOLOv5n, which attains a precision of 93.6%, recall of 89.3%, mAP50 of 95.1%, and mAP50-95 of 72.3%. This indicates enhanced accuracy and robustness in object detection. Similar performance advantages are observed in the small and medium model variants, reinforcing the robustness of the YOLOv11 architecture. Among all the trained models, YOLOv11s demonstrated the highest overall performance, balancing efficiency and detection accuracy.

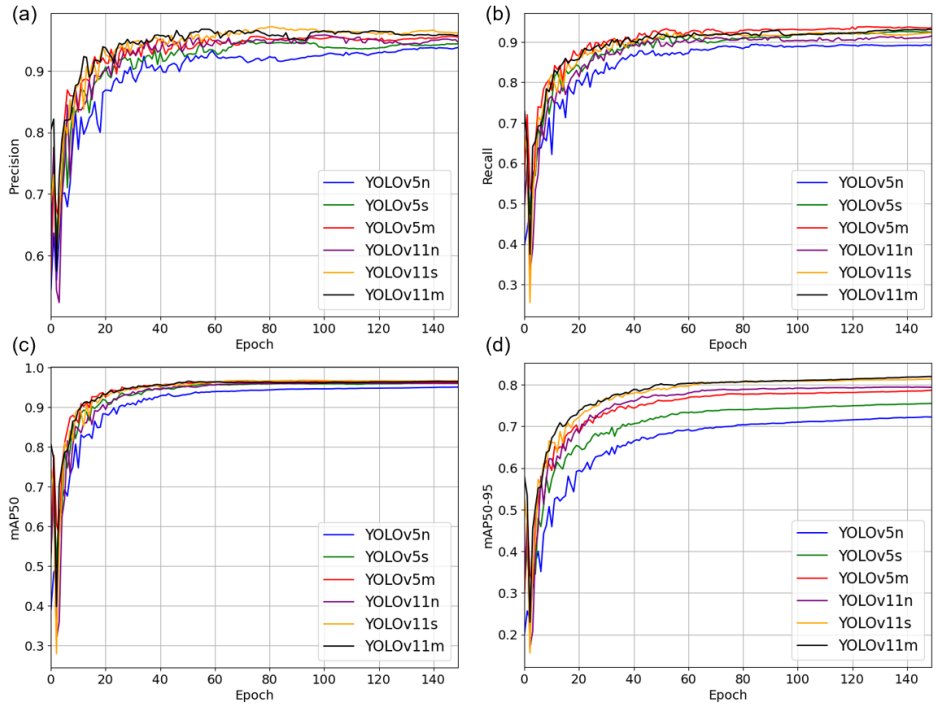

**Fig. 6** Trained results of YOLOv5 and YOLOv11. (a) Training precision of different models; (b) Training recall of different models; (c) The mAP50 scores of different models; (d) The mAP50-95 scores of different models.

Fig. 6 shows the performance and comparative results of various YOLO models. The YOLOv11 models consistently outperform their YOLOv5 counterparts at the same network size, demonstrating superior learning efficiency and higher precision throughout training. The overall training process does not show severe signs of overfitting, as precision continues to improve steadily without abrupt divergence. Larger models achieve higher precision while maintaining stable trends, indicating that the models effectively learn from the dataset.

In summary, based on a comprehensive analysis of evaluation metrics and quantitative performance, the YOLOv11s model was selected as the optimal choice for detecting camera-trapping images.

Model testing

We applied the trained YOLOv11s model to evaluate 1,145 testing images from the Nepal camera trapping dataset with the IoU at 0.5 and a confidence level of 0.5. Testing data was not seen by the model and only used for evaluating model performance. As illustrated in Fig. 7, nearly all the targets were accurately detected, even under challenging conditions such as fast movement (Figs. 7e and 7f), partial body visibility (Figs. 7g and 7h), occlusion by environmental elements (Figs. 7i and 7j), and instances of multiple targets overlapping (Figs 7k and 7l). Additionally, in Figs. 7m and 7n, despite obstructions like a tree affecting the rhino's outline and an anomalously shaped sambar deer climbing a tree, the model successfully identified the targets.

While misclassifications were minimal in the test dataset, certain cases indicate the need for adjusting detection parameters based on different application tasks. For instance, in Fig. 7o, an image containing only the lower legs of a rhino resulted in dual predictions: a rhino at 60% confidence and a human at 52%. On the other hand, in Fig. 7p, two individuals riding bicycles triggered the sensor, with one correctly detected while the other was missing. These instances indicate that while the model exhibits robust classification and detection abilities, refining detection thresholds is still essential for improving its performance in specific applications.

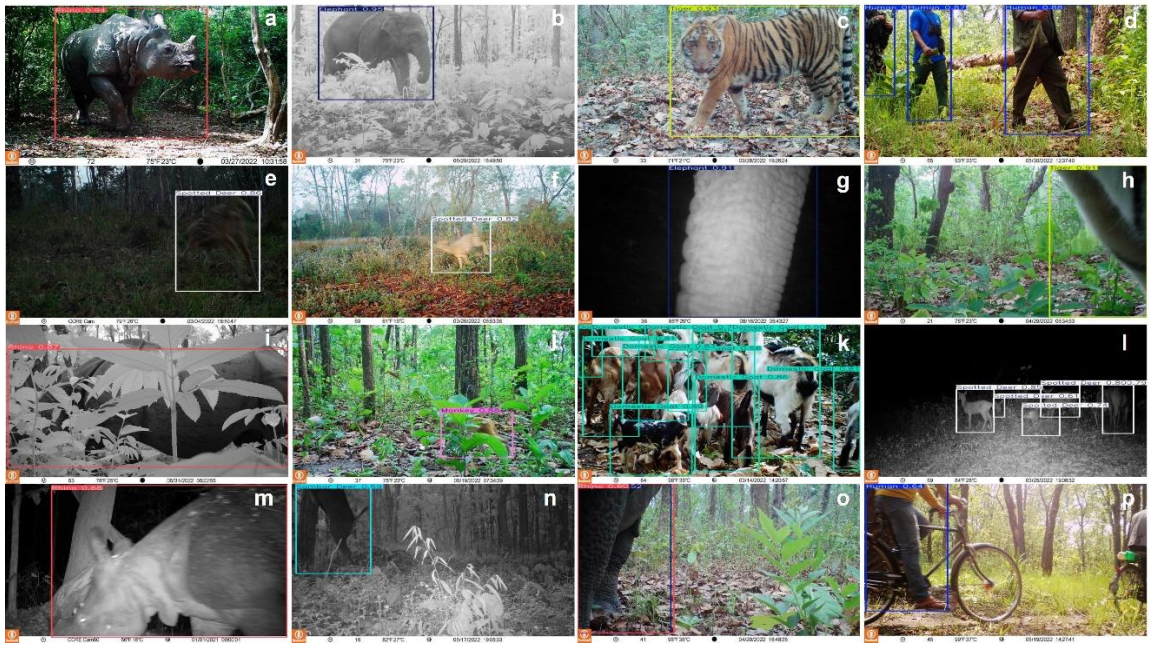

**Fig. 7** Illustration of detection results on the dataset using the YOLOv11s model.

## Hotspot analysis

Utilizing the camera tapping images collected from Chitwan National Park between February and July 2022, we detected all the targets with the YOLOv11s model. As illustrated in Fig. 8, each species displays a distinct pattern of spatial clustering, reflecting differences in habitat use and ecological preferences. Rhinos and elephants are primarily concentrated along the northern boundary of Chitwan National Park (CNP) and within the buffer zone, with rhinos also extending into the corridor forest. This distribution may help explain occasional reports of rhinos entering agricultural areas and damaging crops, resulting in human-rhino conflicts (Bhandari et al., 2022). Tigers show a broader distribution, with detections spanning from the southern to the northern parts of the study area. Deer and rhesus macaques, as the most frequently detected species, were recorded at most camera trap sites, with high-density clusters observed in the corridor forest and northern buffer zone. Notably, deer were not detected within the community forest, suggesting a possible avoidance of areas with high human and livestock presence. A comparison between Figs. 8b and 8e reveals spatial overlap between tiger and deer distributions, supporting the predator-prey relationship between these two species.

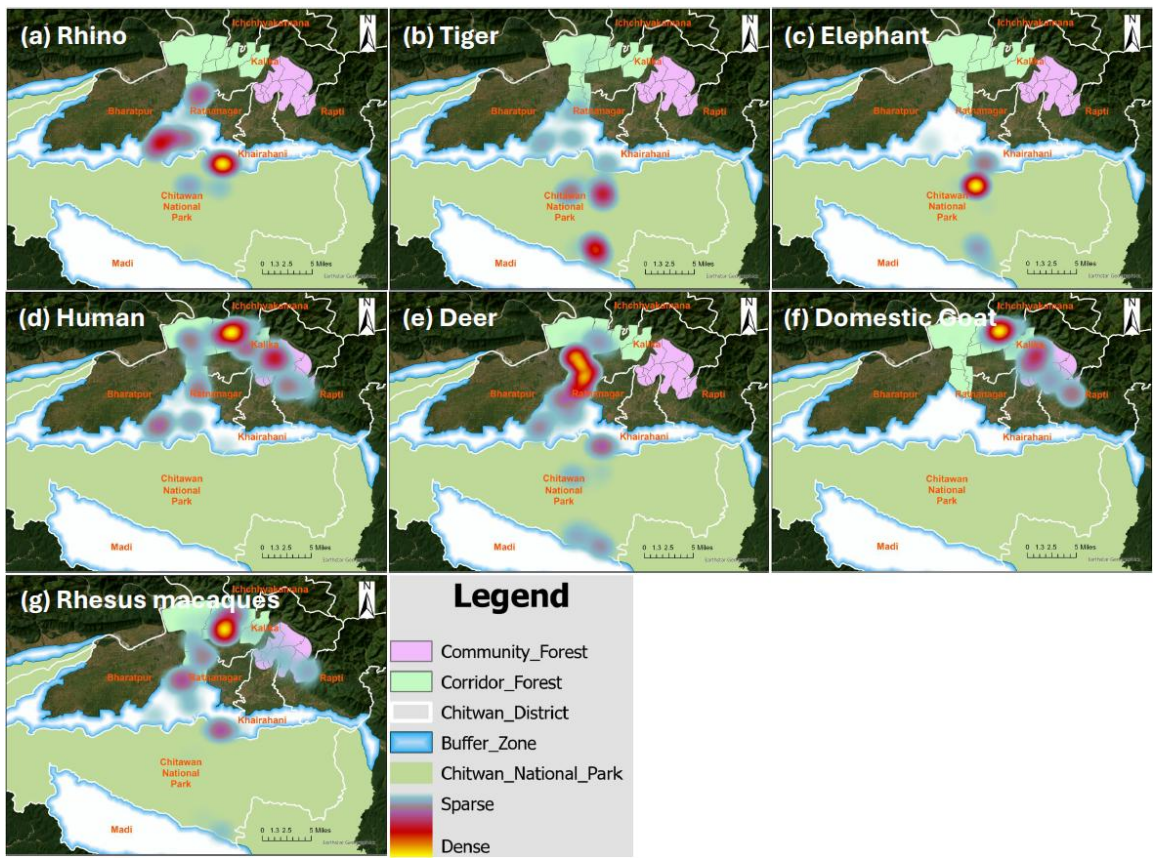

**Fig. 8** Kernel density map of detection outcomes from camera trapping imagery.

Figs. 8d and 8e depict that human activity is most concentrated within the corridor forest, community forest, and northern buffer zone—areas where residents are permitted to collect fuelwood and fodder for subsistence. In contrast, domestic goats are found exclusively in the corridor and community forests, reflecting their close association with human-managed landscapes. The spatial distribution patterns presented in Fig. 8 reveal potential zones of human–wildlife conflict, particularly in transitional landscapes such as the northern buffer zone, corridor forest, and community forest. These areas show strong spatial overlaps between human activity and wildlife presence. The co-occurrence of deer and tigers suggests predator–prey dynamics operating near human settlements, increasing the potential for livestock depredation or human–carnivore encounters. Rhesus macaques clustered near human areas further indicate the likelihood of crop raiding and nuisance behavior.

To further analyze the potential human-wildlife conflicts, we identified human activities, represented by the presence of human and domestic goats, alongside wildlife, including tigers and rhinos, using the trained YOLOv11s model. The detection results consist of image names and labels, combined with GPS information to count the detection frequency of wildlife and human activities captured by each camera trap. Fig. 9 illustrates the hotspots of both selected wildlife and human activities based on the Kernel density analysis of the camera trapping imagery detection results.

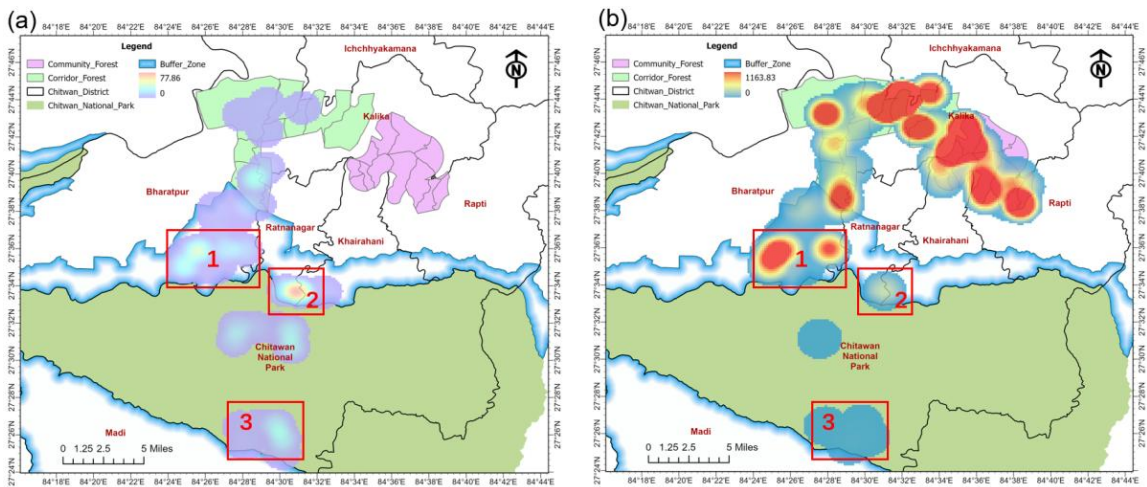

**Fig. 9** Kernel density of (a) wildlife and (b)human activities.

Fig. 9a indicates that wildlife predominantly inhabits areas within the National Park and buffer zones, with some presence extending slightly beyond the buffer zone. In contrast, as shown in Fig. 9b, human activities are concentrated in regions such as Kalika and Rapti, yet there is notable activity within the buffer zones and occasional intrusions into the National Park. Notably, the areas highlighted by rectangles in the figure reveal substantial overlaps between human and wildlife hotspots, pointing to potential zones of human-wildlife conflict.

Clustering results

Based on the detection results and correlation-based hierarchical clustering analysis, we categorized the 80 camera traps into three distinct groups: human, wildlife, and conflict. As illustrated in Fig. 10, camera traps under the human cluster are typically located out of the buffer zone, while those in wildlife clusters are primarily situated within the Chitwan National Park and adjacent buffer zones. Conflict zones are identified where both humans and wildlife are frequently detected by the same camera traps. Buffer zone areas 1 and 2 are notable examples of this, where human intrusion into wildlife territories likely causes disturbance. Additionally, human activities within the National Park at area 3, particularly near the buffer zone, may disrupt natural wildlife inhabits.

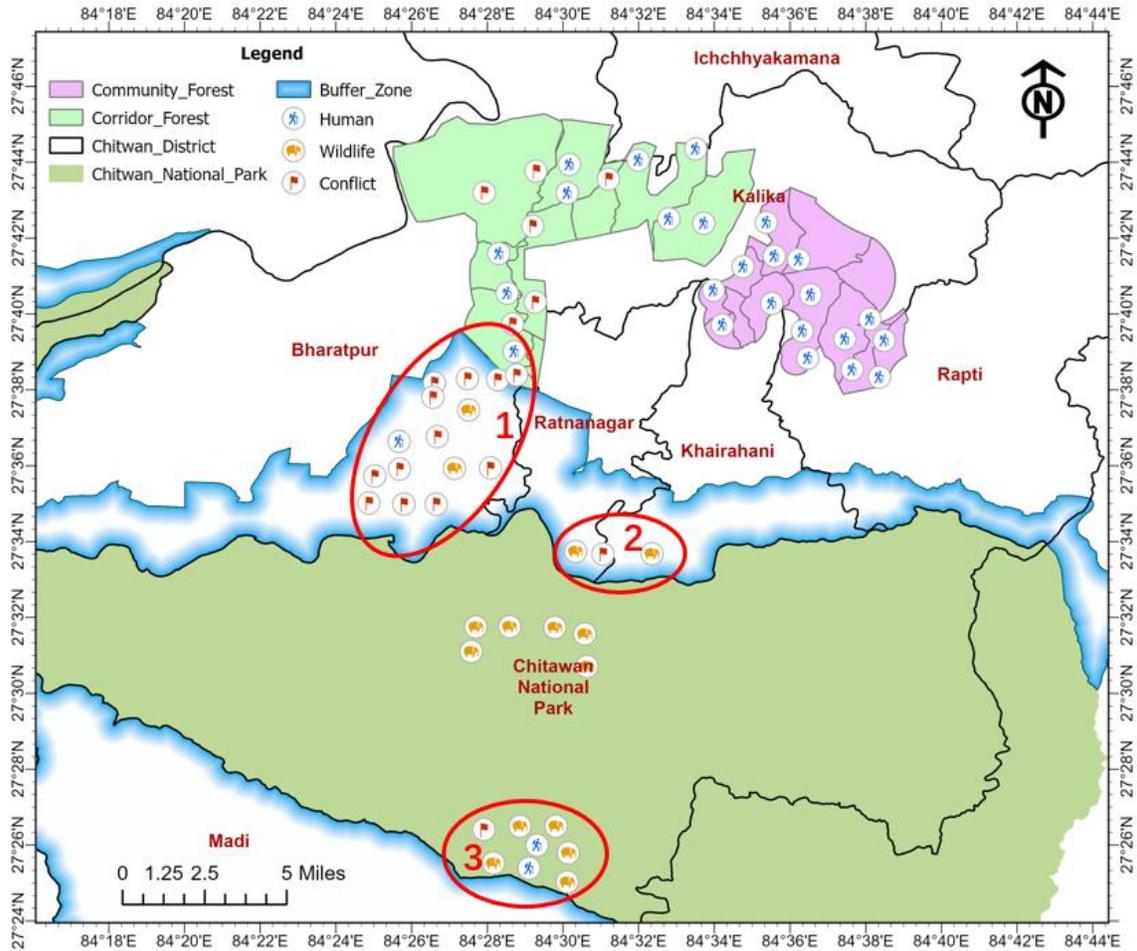

**Fig. 10** Correlation-based hierarchical clustering results.

## Discussion

*Deer survey with drone thermal imagery*

Drone equipped with thermal imaging sensors provided an alternative vantage point for monitoring wildlife and validating detection outcomes derived from camera trap data. The drone surveys were conducted using both manual operation and pre-programmed autonomous flight paths, enabling the monitoring from an overhead perspective. While camera traps offer long-term, passive monitoring from a horizontal ground level with minimal disturbance to wildlife, drones capture observations from a vertical perspective, enhancing spatial coverage and accessibility. These two methods are complementary in nature, and their integration presents a promising framework for future three-dimensional and flexible wildlife monitoring systems.

In this study, we employed an improved Faster R-CNN model to detect deer in the thermal imagery acquired via drone surveys. The detection results generated from this approach are presented in Fig. 11. Through the drone flight path, deer are more concentrated at the northern buffer zone and corridor forest. A comparison of deer detection results from camera traps (Fig. 8e) and drone-based thermal imagery reveals that, despite minor differences in point density intensity, both methods exhibit consistent spatial clustering within the same regions. The observed variation in intensity may be attributed to methodological differences: camera trap detections reflect aggregated detection frequencies over extended deployment periods, whereas drone-derived outputs represent a mosaic of detection events captured during discrete flight sessions. Furthermore, the data were collected during different temporal windows, which may also contribute to variation in detection density. Overall, the deer survey from drone imagery demonstrates the potential for three-dimensional wildlife monitoring and effectively complements and validates the spatial patterns observed through camera trapping.

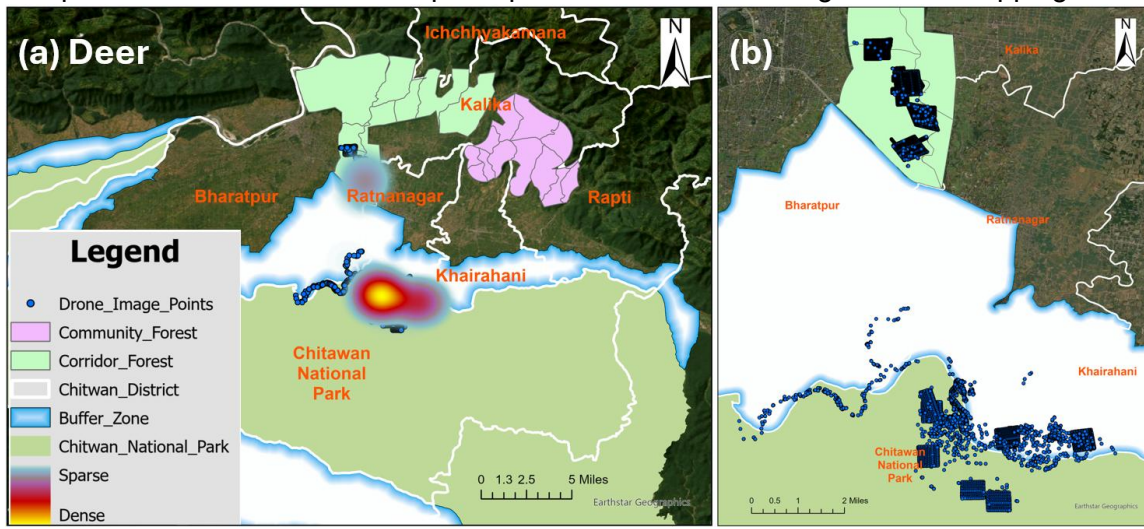

**Fig. 11** Kernel density map of detection outcomes from drone imagery.

*Human-wildlife conflicts*

Identifying human-wildlife conflict is a complex yet crucial task for both human development and wildlife conservation. The camera trap dataset developed in this study proves effective in analyzing such conflicts. The training outcomes highlight the strong performance of the YOLO models, particularly the YOLOv11s. Utilizing the hotspots analysis and hierarchical clustering based on the detection results of the YOLOv11s model, potential conflict zones were identified within the Chitwan National Park and its surrounding buffer zones.

To substantiate these findings, we reviewed the detection data from specific camera traps that indicated possible conflict areas. Fig. 12 illustrates instances where human activities and wildlife presence overlapped at the same locations. As illustrated in Fig. 12a, one individual was observed walking past the camera trap within the Chitwan

National Park where tigers frequently appear. Additionally, within the buffer zone, the situation looks much more frequent. In Fig. 12b, people were observed collecting leaves for livelihood purposes in regions inhabited by wildlife, posing risks to wildlife habitats and potentially leading to injuries. In Fig. 12c, the camera trap captured images of armed individuals, followed by the detection of a rhino just thirty minutes later. This sequence of events raises concerns regarding possible illegal hunting activities, posing a significant threat to wildlife conservation.

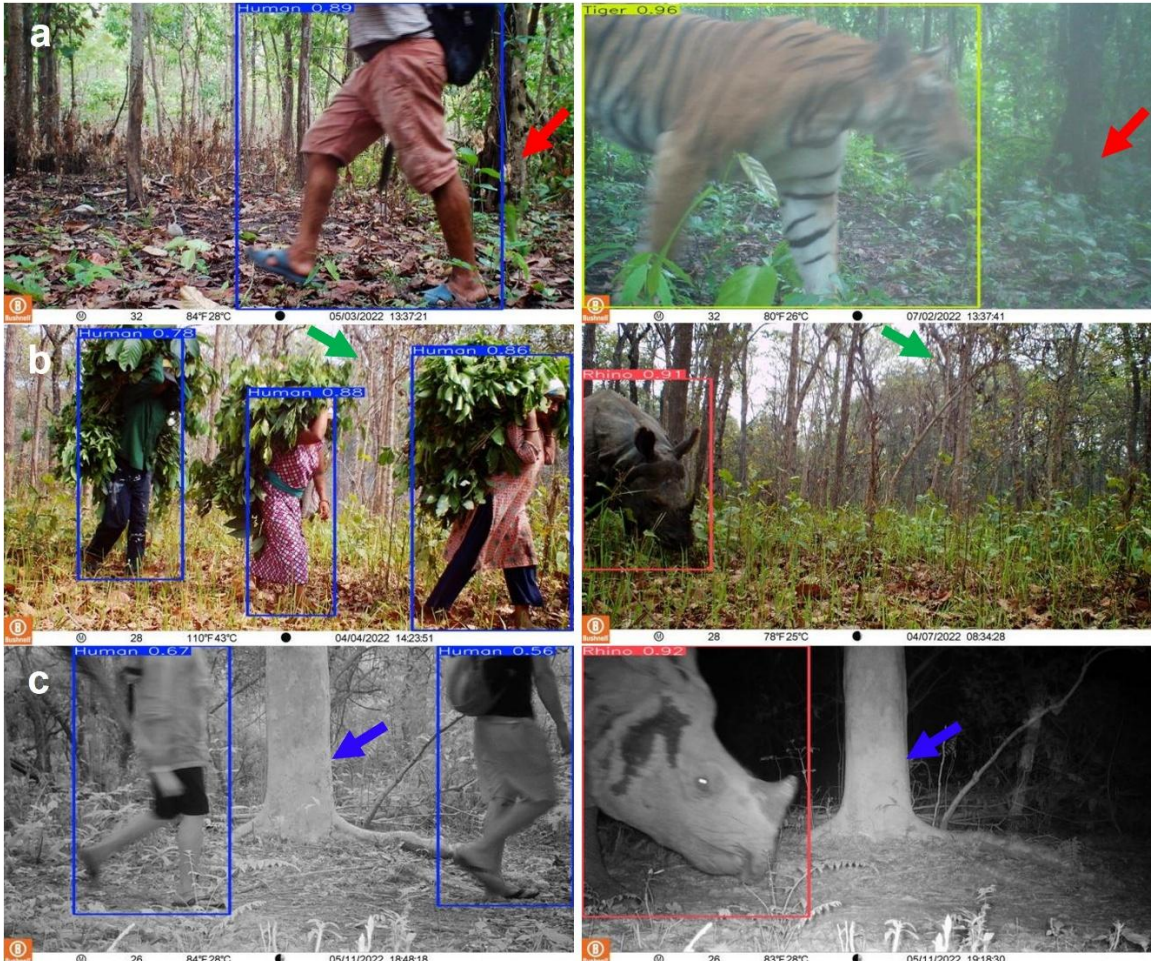

**Fig. 12** Detection results of human activities and wildlife with the identical camera traps of (a) CNP11 (within the Chitwan National Park, area 3), (b) BZF15 (within the buffer zone, area 1), and (c) BZF17 (within the buffer zone, area 2).

## Conclusions

This study demonstrates the effectiveness of integrating multi-scale monitoring approaches with automated object detection by leveraging both camera traps and drone-based data. We introduced a comprehensive dataset derived from camera trap images collected around Chitwan National Park, Nepal. Through the training and evaluation of YOLOv5 and YOLOv11 models across six different network architectures, the deep learning models exhibited high accuracy and robustness in detecting both wildlife and human presence. The resulting detections were subjected to spatial analyses, including hotspot mapping and hierarchical clustering, to identify potential areas of human–wildlife conflicts.

In addition, thermal imagery collected via drone was detected using an enhanced Faster R-CNN model, contributing to a complementary vertical perspective that supports three-dimensional wildlife monitoring and cross-validates camera trap detections. These findings offer valuable insights into conservation planning, providing actionable data to inform local authorities in the development of targeted management strategies. Future research will aim to advance this framework by incorporating automated individual identification and analyzing spatial interactions using state-of-the-art machine learning and geographic information system (GIS) techniques within protected landscapes.

## CRediT authorship contribution statement

**Hao Chen:** Writing – original draft, Writing – review & editing, Data curation, Visualization, Validation, Software, Methodology, Investigation. **Fang Qiu**: Writing – review & editing, Supervision, Project administration, Funding acquisition, Conceptualization. **Li An**: Conceptualization, Project administration, Funding acquisition. **Eve Bohnett**: Data curation, Formal analysis, methodology. **Haitao Lyu**: Data curation, Visualization, Validation. **Shuang Tian**: Data curation, Conceptualization.

## Declaration of competing interest

The authors declare that they have no known competing financial interests or personal relationships that could have appeared to influence the work reported in this paper.

## Acknowledgments

This research was performed with financial support from the National Science Foundation (NSF) of the USA grant number BCS-1826839. We thank the Chitwan National Park, notably the protected area managers and staff who supported the field work.

## Data availability

The datasets generated during and/or analyzed during the current study are available from the corresponding author upon reasonable request.